\documentclass[11pt]{article}

\usepackage[]{acl}
\usepackage{multirow}
\usepackage{times}
\usepackage{latexsym}
\usepackage{rotating}
\usepackage{array}
\usepackage[T1]{fontenc}

\usepackage[utf8]{inputenc}

\usepackage{microtype}

\usepackage{inconsolata}

\usepackage{graphicx}

\title{What is lost in Normalization? Exploring Pitfalls in Multilingual ASR Model Evaluations}

\author{
 \textbf{Kavya Manohar\textsuperscript{1,2}}
 \textbf{Leena G Pillai\textsuperscript{1,3}}
  \textbf{Elizabeth Sherly\textsuperscript{1}}
\\
 \\
 \textsuperscript{1}Digital University Kerala\\
 \textsuperscript{2}Swathanthra Malayalam Computing
 \textsuperscript{3}University of Kerala
\\
 \small{
   \textbf{Correspondence:} \href{kavya.manohar@duk.ac.in}{kavya.manohar@duk.ac.in}, \href{leena.g@duk.ac.in}{leena.g@duk.ac.in}
 }
}

\begin{document}
\maketitle
\begin{abstract}
This paper explores the pitfalls in evaluating multilingual automatic speech recognition (ASR) models, with a particular focus on Indic language scripts. We investigate the text normalization routine employed by leading ASR models, including OpenAI Whisper, Meta's MMS, Seamless, and Assembly AI's Conformer, and their unintended consequences on performance metrics. Our research reveals that current text normalization practices, while aiming to standardize ASR outputs for fair comparison, by removing inconsistencies such as variations in spelling, punctuation, and special characters, are fundamentally flawed when applied to Indic scripts. Through empirical analysis using text similarity scores and in-depth linguistic examination, we demonstrate that these flaws lead to artificially improved performance metrics for Indic languages. We conclude by proposing a shift towards developing text normalization routines that leverage native linguistic expertise, ensuring more robust and accurate evaluations of multilingual ASR models. 
\end{abstract}

\section{Introduction}
Automatic speech recognition (ASR) systems have become increasingly relevant in various applications, ranging from voice assistants and transcription services to accessibility tools for the disabled population. The performance and usability of ASR models are evaluated in terms of their error rates. Recent advancements in open ASR models pretrained in self-supervised \cite{schneider19_interspeech,babu22_interspeech, chung2021w2v} manner or weakly supervised \cite{radford2023robust} manner are capable of handling various languages and scripts. These models can be fine-tuned for improved performance in domains or languages of interest. This capability has revolutionized speech recognition in ultra low resource languages and scenarios \cite{rouditchenko23_interspeech}. Many of these models have brought down state of the art (SOTA) word error rates (WERs) on popular benchmarks.

Evaluation of the performance of ASR models are often affected by the prediction differing from the ground truth in letter casing, punctuation, spelling variants etc. leading to inflated WERs. To mitigate this, a text normalization routine is employed \cite{deviyani22_s4sg, zhang21ja_interspeech}. A proper text normalization routine is required to minimize penalization of non-semantic differences by aligning the predicted output more closely with the ground truth. 

The study presented in this paper examines the pitfalls in the current normalizations routines employed in the latest ASR models on the banchmarking of non-English languages, specifically on many Asian languages that use Indic scripts\footnote{\url{https://en.wikipedia.org/wiki/Brahmic_scripts}}. Our empirical analysis reveals that the current normalization practices can result in significant errors, particularly in many low-resource languages, by boosting the model performance on many benchmarks and misleading the research community.  We propose for the development of linguistically informed normalization routines that account for the unique characteristics of each language, ensuring a fair and reasonable evaluation and benchmarking process for multilingual ASRs.

\vspace{-0.2cm}

\begin{table*}[!h]
    \centering
    \begin{tabular}{c}
         \includegraphics[width=1\textwidth]{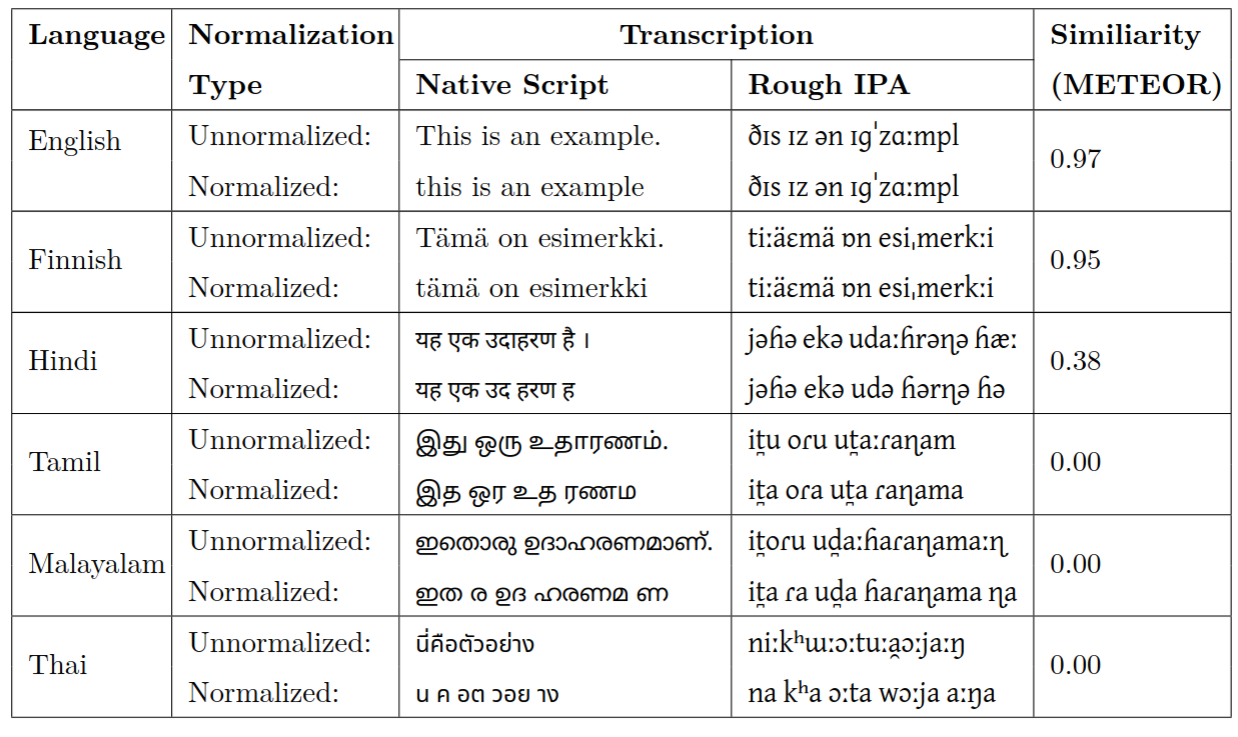}
    \end{tabular}
    \caption{A demonstration of the effect of Whisper normalization. While diacritics are retained in non-English languages (eg: Finnish) that uses latin script, the relevant vowel signs and virama sign are lost in Indic scripts. Rough Romanized transcript in IPA is also provided. Text similarity between original and Whisper-normalized text are indicated using METEOR score \cite{banerjee-lavie-2005-meteor}}
    \label{tab:normalization}
\end{table*}

\section{Background and Related Works}

Prior to the introduction of OpenAI's Whisper model \cite{radford2023robust}, most ASR systems were trained on normalized text transcripts and produced output without punctuation or casing. Whisper, however, outputs UTF-8 text, requiring a comprehensive normalization process to accurately evaluate its performance. This ensures that the evaluation metric, WER penalizes only actual word mis-transcriptions, not formatting or punctuation differences.

Whisper's normalization routine for English extends beyond basic casing and punctuation, incorporating transformations such as converting contracted abbreviations to expanded forms and expanding currency symbols. However, this approach would require a language-specific set of transformations for non-English text. Due to the lack of linguistic knowledge to develop such normalizers for all languages, the Whisper's normalization relies on a basic data-driven approach, which includes replacement of characters in the \texttt{mark} class with spaces and removes successive whitespace characters to a single instance \cite{radford2023robust}.

The non-English normalization routine employed by Whisper, inadvertently removes vowel signs (\textit{matras}), that belong to the the \texttt{mark} class of Unicode characters. These vowel signs, essential for correct word formation and pronunciation, are removed along with other punctuation marks, leading to significant distortions in the text in languages such as Hindi, Bengali, Tamil, and others \cite{Ross2023, Manohar2024}. This results in words being broken down into consonants without their associated vowels, causing a loss of meaning and intelligibility. This also leads to incorrect WER calculations for languages written in Indic scripts. Additionally, Thai, which does not use spaces between words but relies on spaces to delimit sentences, is also affected. The normalization process inserts spaces instead of vowel signs, effectively distorting the nature of the language. See Table \ref{tab:normalization} for examples with detailed analysis provided in section \ref{intrinsic}.

This normalization routine has been adopted by various later models, including Meta's MMS, Seamless series \cite{pratap2024scaling, barrault2023seamlessm4t, barrault2023seamless} and AssemblyAI's Conformer-1 \cite{conformer1} for evaluation and benchmarking and is integrated into Huggingface transformers\footnote{\href{https://github.com/huggingface/transformers/blob/main/src/transformers/models/whisper/english_normalizer.py}{Normalization in Huggingface Whisper Transformer}}, thus amplifying its impact.


\section{Methodology}

In this study, we present two complementary empirical evaluations to assess the impact of Whisper's normalization routine on different languages. First, we conduct an intrinsic evaluation by comparing the similarity of example sentences from various languages before and after normalization, using the METEOR score as a text similarity metric. Second, we perform an extrinsic evaluation by measuring the WER on a multilingual benchmark dataset for the same set of languages, with and without the application of Whisper's normalization. For our case study, we used both the baseline and fine-tuned Whisper ASR because their outputs include punctuation, unlike other ASR models in the literature. This allowed us to demonstrate the impact of normalization on ASR outputs with punctuation.  All the datasets and the models used in this experiments are available under permissive licenses in Huggingface repositories and listed in Appendix \ref{sec:appendix}. All the evaluations were run on a single NVIDIA A100 GPU.

\subsection{Analysis of Text Similarity after Whisper Normalization}
\label{intrinsic}
\begin{figure*}[htpb]
  \includegraphics[width=\textwidth]{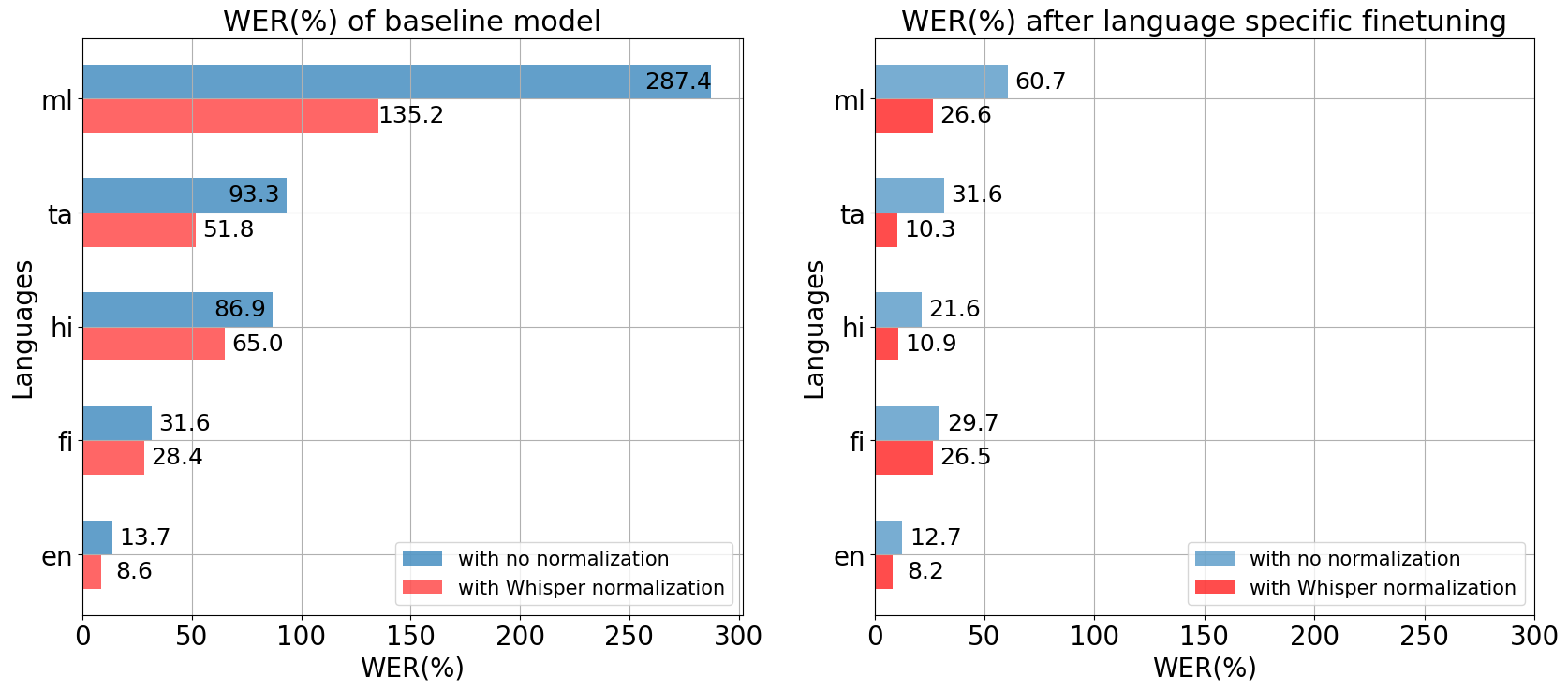}
  \caption{Performance comparison of the OpenAI Whisper-Small model across different languages. The graph on left shows WER on the original model and the one on right shows the result after language specific finetuning. Regular WER are computed on raw transcripts and normalized WER are computed on Whisper normalized transcripts.}
  \label{fig:wer}
  \vspace{-0.2cm}
\end{figure*}

To empirically assess the impact of Whisper's normalization routine on different languages, we conducted a comparative analysis of example sentences from languages that employ various script systems. Specifically, we selected languages that use Latin script (English and Finnish), Indic scripts (Hindi, Tamil, and Malayalam), and South East Asian scripts (Thai). For each language, we prepared a set of example sentences that were identical in meaning but differed in their script and formatting as presented in Table \ref{normalization analysis}. The METEOR score \cite{banerjee-lavie-2005-meteor} was employed to quantify the similarity between the original and normalized sentences. It is a text similarity metric that considers the precision, recall, and F-score of the machine-translated text, providing a comprehensive measure of its similarity to the reference text, while also placing importance on the order of words in the text.

The similarity scores we obtained demonstrate the varying impact of Whisper's normalization routine on different languages. The high similarity scores for English (0.97) and Finnish (0.95) indicate that the normalization process preserves the linguistic structure and meaning of these languages very well. The diacritic marks in Finnish are retained without any distortion as indicated in the Table \ref{normalization analysis}. This is because the normalization routine ensures the diacritic marks gets converted to \texttt{letter} class of characters using NFKC compatibility composition rules of Unicode\footnote{\url{http://unicode.org/reports/tr15/}}, before \texttt{mark} class of characters are replaced by space.

In contrast, as illustrated in Table \ref{normalization analysis}, the normalization process severely distorts the text in languages other than English and Finnish. The replacement of Unicode characters in the \texttt{mark} class, including vowel signs and \textit{virama} symbols, by spaces after Whisper normalization significantly alters the linguistic structure of these languages. While Hindi, with a METEOR score of 0.38, is less affected due to its analytic typology, Malayalam and Tamil are severely impacted \cite{bharadwaja2007statistical, manohar2020quantitative} by the splitting of morphologically complex words at every occurrence of vowel signs and \textit{virama} symbols, leading to similarity scores of 0. Thai, which typically does not use spaces between words, is also affected by the removal of important vowel signs, resulting in a text that is unusable due to excessive spacing and a similarity  score of 0.

\subsection{Impact of Whisper Normalization on WER}

To empirically analyze the impact of normalization on the WER, we present the results of evaluating the original Whisper-small model, referred to as the baseline model, with and without the application of Whisper's normalization on the  test split of Google FLEURS \cite{fleurs2022arxiv} multilingual speech dataset.

The left side bar graph in Figure \ref{fig:wer} shows that the WER  of the baseline model is significantly high for languages other than English and Finnish, with values of 86.9\% for Hindi, 93.3\% for Tamil, and 287.4\% for Malayalam. The baseline ASR model exhibits a WER exceeding 100\% for Malayalam due to a high number of insertion errors, leading to the combined total of substitutions, deletions, and insertions surpassing the total word count in the reference transcript. While the application of Whisper's normalization results in modest WER improvements for English and Finnish, with an absolute reduction of 5.1\% and 3.2\% respectively, Indic languages experience suspicious absolute WER reductions: 21.9\% for Hindi, 41.5\% for Tamil, and a substantial 152.2\% for Malayalam. 


Due to the poor performance of the baseline model on many Indian languages, we conducted a further comparison of WER with and without Whisper's normalization on publicly available models that have been derived from the baseline model after language-specific fine-tuning. The fine-tuned models used in these evaluations are listed in Appendix \ref{sec:appendix}. Fine-tuning has significantly improved the performance of the Hindi, Tamil, and Malayalam models. 

Fine-tuned models of English and Finnish exhibit a reasonable absolute reduction of 4.5\% and 3.2\% on WER respectively. In contrast, Indic languages exhibit a substantial absolute reduction in WER, with decreases of 10.7\% for Hindi, 21.3\% for Tamil, and 34.1\% for Malayalam. Notably, the languages that showed the worst similarity scores exhibit the maximum improvement in WER after normalization. This suggests that the normalization process, which breaks most words into a series of consonants and adds spaces, artificially increases the number of words in the reference, thereby reducing the WER.

\section{Recommendations}

Findings from our empirical evaluation underscore the importance of language-specific normalization routines to ensure accurate text representation and reliable performance evaluation in many underrepresented Indic languages. Building up on our findings, we propose a collaborative approach, leveraging the collective efforts of native speakers and linguistic experts to develop effective normalization routines for diverse linguistic contexts.




\section{Conclusions}

The empirical evaluation conducted in this study highlights that the current practice of normalization severely affects the text representation across languages, resulting in artificially boosted WER and SOTA performance. By adopting a more tailored approach to evaluations, we can enhance the reliability of multilingual ASR models, making them truly inclusive and effective across diverse linguistic landscapes.

\section{Limitations}

\begin{enumerate}
    \item Being a position paper, this study highlights only the limitations of existing normalization techniques, but does not propose new normalization algorithms. 
    \item The results are based on specific datasets  and publicly available models used for evaluating WER. Variability in datasets (e.g., different accents, dialects, or recording conditions) might influence the reported values.
    \item The primary metric discussed is WER. Other evaluation metrics (e.g., phoneme error rate, semantic error rate, match error rate) might provide additional insights into the impacts of text normalization.
    \item We used the \texttt{raw transcription} field of the FLEURS corpus, which could be a reason for the difference from the WER values reported in \citet{radford2023robust}.
    \item While the paper focuses on text normalization on Indian languages there could be other languages which gets affected by the normalization differently.
    \item We omitted Thai from WER comparison charts because for languages where space is not a word delimiter, character error rate is the metric reported in \citet{radford2023robust}.
\end{enumerate}

\bibliography{acl_latex}

\appendix

\section{Resources}
\label{sec:appendix}

We have used the following publicly available models and datasets for our experiments.

\subsection*{ASR Models}

\begin{enumerate}
    \item The baseline model:\\ \url{https://huggingface.co/openai/whisper-small}
    \item The Fine-tuned English:\\ \url{https://huggingface.co/openai/whisper-small.en}
    \item The Fine-tuned Finnish:\\ \url{RASMUS/whisper-small-fi-15k_sample}
    \item The Fine-tuned Hindi:\\ \url{https://huggingface.co/vasista22/whisper-hindi-small}
    \item The Fine-tuned Tamil:\\ \url{https://huggingface.co/vasista22/whisper-tamil-small}
    \item The Fine-tuned Malayalam:\\ \url{https://huggingface.co/vrclc/Whisper_small_malayalam}
\end{enumerate}

\subsection*{Speech Dataset}

\begin{enumerate}
    \item Google FLEURS:\\ \url{https://huggingface.co/datasets/google/fleurs}

\end{enumerate}

\end{document}